\documentclass[conference]{IEEEtran}
\IEEEoverridecommandlockouts
\usepackage{cite}
\usepackage{amsmath,amssymb,amsfonts}
\usepackage{algorithmic}
\usepackage{graphicx}
\usepackage{textcomp}
\usepackage{xcolor}
\usepackage{makecell}
\usepackage{multirow} 

\usepackage[section]{placeins}

\def\BibTeX{{\rm B\kern-.05em{\sc i\kern-.025em b}\kern-.08em
    T\kern-.1667em\lower.7ex\hbox{E}\kern-.125emX}}
\begin{document}

\title{SCB-Dataset3: A Benchmark for Detecting Student Classroom Behavior\\

}

\author{\IEEEauthorblockN{1\textsuperscript{st} YANG FAN}
\IEEEauthorblockA{\textit{Jinan University} \\
Guangzhou, China \\
winstonyf@qq.com}
\and
\IEEEauthorblockN{2\textsuperscript{nd} WANG TAO}
\IEEEauthorblockA{\textit{Chengdu Neusoft University} \\
Chengdu, China \\
wang-tao@nsu.edu.cn}
}

\maketitle

\begin{abstract}
The use of deep learning methods to automatically detect students' classroom behavior is a promising approach for analyzing their class performance and improving teaching effectiveness. However, the lack of publicly available datasets on student behavior poses a challenge for researchers in this field. To address this issue, we propose the Student Classroom Behavior dataset (SCB-dataset3), which represents real-life scenarios. Our dataset comprises 5686 images with 45578 labels, focusing on six behaviors: hand-raising, reading, writing, using a phone, bowing the head, and leaning over the table. We evaluated the dataset using the YOLOv5, YOLOv7, and YOLOv8 algorithms, achieving a mean average precision (map) of up to 80.3$\%$. We believe that our dataset can serve as a robust foundation for future research in student behavior detection and contribute to advancements in this field. Our SCB-dataset3 is available for download at: https://github.com/Whiffe/SCB-dataset

\end{abstract}


\section{Introduction}

\begin{table*}
    \centering \renewcommand\arraystretch{1.25} 
    \caption{Datasets and methods of student classroom behavior in recent years}
    \label{Datasets_Methods}
    \begin{tabular}{p{2.2cm}p{6.5cm}p{7.6cm}}
        \hline
         Dataset & Content  & Method \\ \hline
         ActRec-Classroom\cite{Fu_behavior_analysis} &  The dataset comprises of 5 categories: listening, fatigue, raising hand, leaning and reading/writing with 5,126 images. & First, Faster R-CNN detects human bodies. Then, OpenPose extracts key points of skeletons, faces, and fingers. Finally, a CNN-based classifier is developed for action recognition. \\
         A large-scale dataset for student behavior \cite{Zheng_student_behavior} &  The dastaset from thirty schools, labeling these behaviors using bounding boxes frame-by-frame, which contains 70k hand-raising samples, 20k standing samples, and 3k sleeping samples.& An enhanced Faster R-CNN model for student behavior analysis, which includes a scale-aware detection head, a feature fusion strategy, and the utilization of OHEM for improved detection performance with reduced computation overhead and class imbalances in a real corpus dataset. \\
        BNU-LCSAD \cite{Sun_Student_Class_Behavior_Dataset} & A comprehensive dataset that can be employed for recognizing, detecting, and captioning students’ behaviors in a classroom. Dataset include 128 videos in different disciplines and in 11 classrooms.    & The baseline models used for different tasks include the two-stream network for recognition, ACAM\cite{ACAM} and MOC\cite{MOC} for action detection, BSN\cite{Bsn} and DBG\cite{DBG} for temporal detection, and RecNet\cite{RecNet} and HACA\cite{HACA} for video captioning. \\
        Student Classroom Behavior Dataset\cite{Zhou_Classroom_Learning}  & The dataset included 400 students from 90 classroom videos in a primary school. The images were collected using single-person images. There are 10,000 images of students raising their hands, walking back and forth, writing on the blackboard, and looking up and down. Additionally, there are 1,000 images of students bending down, standing, and lying on the table. & The proposed method utilizes a 10-layer deep convolutional neural network (CNN-10) to recognize student classroom behaviors by extracting key information from human skeleton data, effectively excluding irrelevant information, and achieving higher recognition accuracy and generalization ability.  \\
        Student behavior dataset\cite{Li_Student_behavior} & A dataset of student behavior based on an intelligent classroom was established.   Student behavior was divided into seven classes and challenging class surveillance videos were selected for annotation to generate the dataset. & The proposed method integrates relational features to analyze how actors interact with their surrounding context.  It models human-to-human relationships using body parts and context, and combines these relational features with appearance features for accurate human-to-human interaction recognition.  \\
        Student action dataset\cite{Trabelsi_Classroom} & A dataset that contains 3881 labeled images of various student behaviors in a classroom environment, including raising hands, paying attention, eating, being distracted, reading books, using a phone, writing, feeling bored, and laughing.   & The YOLOv5 object detection model is used to train and evaluate . \\
        A large-scale student behavior dataset\cite{Zhou_Stuart} & The dataset that contains five representative student behaviors highly correlated with student engagement (raising hand, standing up, sleeping, yawning, and smiling) and tracks the change trends of these behaviors during the course. & The proposed StuArt, an innovative automated system that enables instructors to closely monitor the learning progress of each student in the classroom. StuArt also includes user-friendly visualizations to facilitate instructors' understanding of individual and overall learning status.\\
        Classroom behavior dataset\cite{BiTNet} & The dataset was collected from publicly accessible education videos and contains genuine images.  These videos cover primary, middle, and university levels.  After classification and annotation, the dataset consists of 4432 images and 151574 annotation boxes.  the dataset contains 7 typical student behaviors (writing, reading, listening, raising hand, turning around, standing, and discussing) and 1 typical teacher behavior (guiding). & Authors propose BiTNet, a real-time object detection network, for enhancing teaching quality and providing feedback through real-time analysis of student behavior in classroom.   BiTNet addresses challenges faced by current methods, such as occlusion and small objects in images.  \\
        Teacher Behavior Dataset (STBD-08) \cite{CBPH-Net} & The dataset contains 4432 images with 151574 labeled anchors covering eight typical classroom behaviors. &  To address these challenges such as occlusions, pose variations, and inconsistent target scales, authors propose an advanced single-stage object detector called ConvNeXt Block Prediction Head Network (CBPH-Net).    \\ 

        \hline
    \end{tabular}
\end{table*}

In recent years, with the development of behavior detection technology\cite{Zhu_comprehensive_study}, it has become possible to analyze student behavior in class videos to obtain information on their classroom status and learning performance. This technology is of great importance to teachers, administrators, students, and parents in schools. However, in traditional teaching models, teachers find it difficult to pay attention to the learning situation of every student and can only understand the effectiveness of their own teaching methods by observing a few students. School administrators rely on on-site observations and student performance reports to identify problems in education and teaching. Parents can only understand their child's learning situation through communication with teachers and students. Therefore, utilizing behavior detection technology to accurately detect student behavior and analyze their learning status and performance can provide more comprehensive and accurate feedback for education and teaching.

Existing student classroom behavior detection algorithms can be roughly divided into three categories: video-action-recognition-based\cite{HUANG_Multi-person_classroom_action}, pose-estimation-based\cite{He_recognition_student_classroom_behavior} and object-detection-based\cite{YAN_Student_classroom_behavior}. Video-based student classroom behavior detection enables the recognition of continuous behavior, which requires labeling a large number of samples. For example, the AVA dataset\cite{Ava} for SlowFast\cite{Slowfast} detection is annotated with 1.58M. And, video behavior recognition detection is not yet mature, as in UCF101\cite{UCF101} and Kinetics400\cite{kinetics}, some actions can sometimes be determined by the context or scene alone. Pose-estimation-based algorithms characterize human behavior by obtaining position and motion information of each joint in the body, but they are not applicable for behavior detection in overcrowded classrooms. Considering the challenges at hand, object-detection-based algorithms present a promising solution. In fact, in recent years object-detection-based algorithms have made tremendous breakthroughs, such as YOLOv5\cite{YOLOv5}, YOLOv7\cite{YOLOv7}, YOLOv8\cite{YOLOv8}. Therefore, we have employed an object-detection-based algorithm in this paper to analyze student behavior.

In this study, we have iteratively optimized our previous work to further expand the SCB-Dataset\cite{SCB-Dataset1,SCB-Dataset2}.  Initially, we focused solely on the behavior of students raising their hands, but now we have expanded to include six behaviors: hand-raising, reading, writing, using phone, bowing the head, leaning over the table.  Through this work, we have further addressed the research gap in detecting student behaviors in classroom teaching scenarios.

We have conducted extensive data statistics and benchmark tests to ensure the quality of the dataset, providing reliable training data.

Our main contributions are as follows:

1. We have updated the SCB-Dataset to its third version (SCB-Dataset3), with the addition of 6 behavior categories. This dataset contains a total of 5686 images with 45578 annotations. It covers different scenarios from kindergarten to university.

2. We conducted extensive benchmark testing on the SCB-Dataset3, providing a solid foundation for future research.

3. For the university scene data in SCB-Dataset3, we employed the "frame interpolation" method and conducted experimental validation. The results show that this method significantly improves the accuracy of behavior detection.

4. We proposed a new metric, the Behavior Similarity Index (BSI), which measures the similarity in form between different behaviors under a network model.

\section{Related Works}
\noindent\textbf{Student classroom behavior datasets and methods}


Recently, many researchers have utilized computer vision to detect student classroom behaviors. However, the lack of public student behavior datasets in the education field restricts the research and application of behavior detection in classroom scenes. Many researchers have also proposed many unpublished datasets, the relevant datasets and methods are shown in Table \ref{Datasets_Methods}.

\section{SCB-Dataset}

Understanding student behavior is crucial for comprehending their learning process, personality, and psychological traits, and is important in evaluating the quality of education.       The hand-raising, reading, writing, using phone, bowing the head, leaning over the table behaviors are important indicators of evaluating classroom quality.     However, the lack of public datasets poses a significant challenge for AI research in the field of education.   Based on our previous work, we have further enhanced the SCB-Dataset by increasing the number of samples significantly, expanding the student behavior classes from 3 to 6, and introducing university classroom scenarios.   This improved dataset is now referred to as SCB-Dataset3. 

 For the classroom dataset from kindergarten to high school (which we call SCB3-Dataset-S). We collected over a thousand videos ranging from kindergarten to high school, each with a duration of approximately 40 minutes. To reduce imbalances in behavior classes and improve representativeness and authenticity, we intentionally selected 3 to 15 frames with a specific time interval for each video. The videos were extracted from the "bjyhjy", "1s1k", "youke".  "qlteacher" and "youke-smile.shec" websites.

\begin{figure}[htbp]
\centerline{\includegraphics[width=0.48\textwidth]{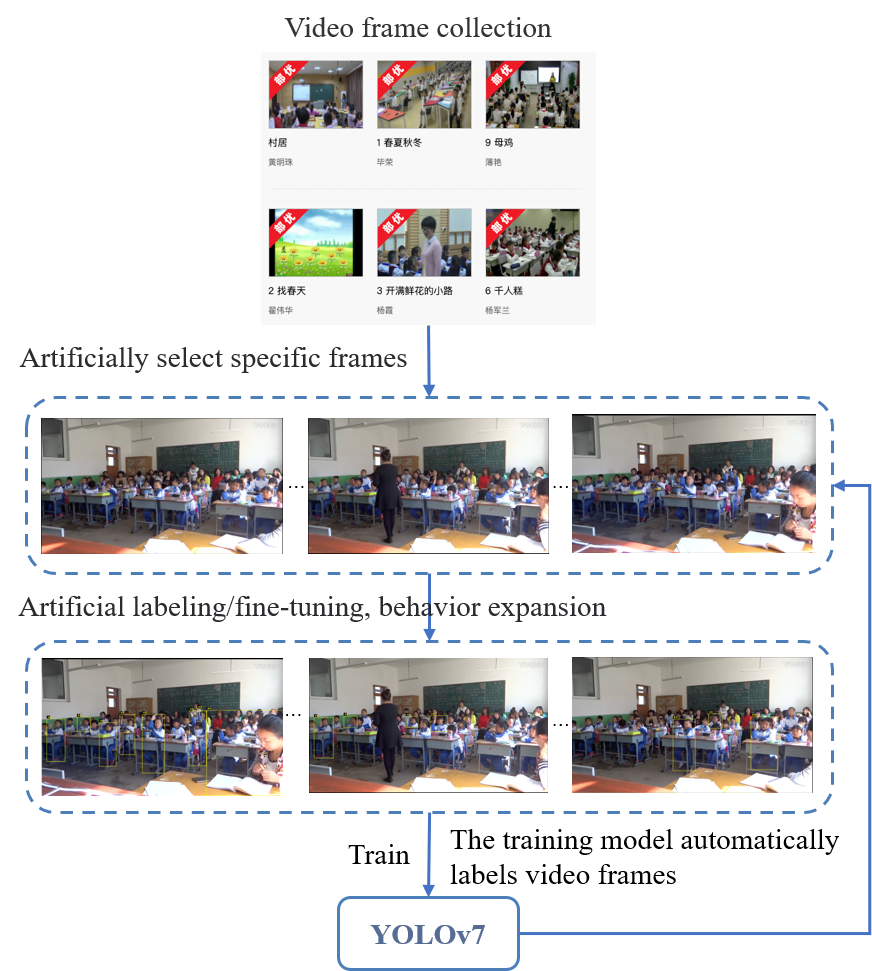}}
\caption{Process for creating SCB-Dataset3-S.}
\label{Process-of-SCB3-S}
\end{figure}

 Our SCB-Dataset3-S process, as shown in Fig.~\ref{Process-of-SCB3-S}, adopts an iterative training approach. Firstly, we collected a series of suitable video frames from an online course website and carefully screened them. Next, we manually annotated these video frames to provide accurate label data for subsequent training.

Then, we input the annotated data into the YOLOv7 model for training. By using this model, we are able to perform inference detection on the unannotated video frames, i.e., the model detects the target objects in the videos. We further fine-tune these detection results using artificial intelligence techniques to improve the accuracy and reliability of the detection while reducing human effort.

Finally, we input the fine-tuned data back into the YOLOv7 model for training again, in order to further optimize the model's performance. Through iterative training, we continuously improve the accuracy and robustness of the model, resulting in more accurate detection results.

\begin{figure}[htbp]
\centerline{\includegraphics[width=0.48\textwidth]{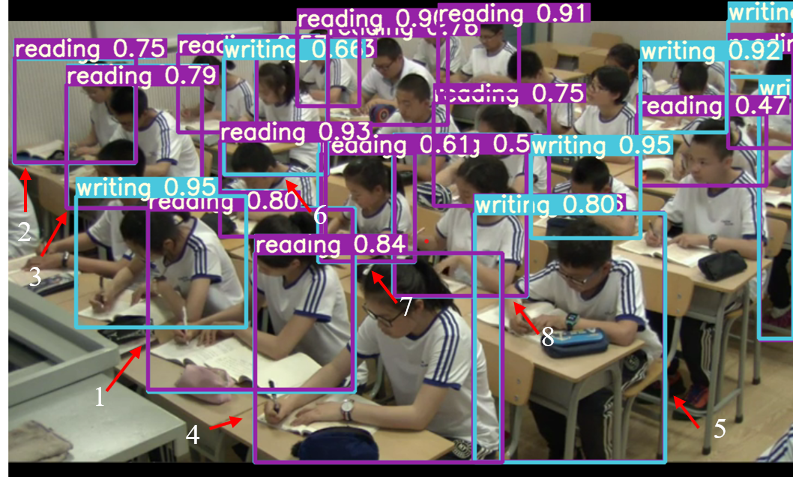}}
\caption{ Overlapping results of detection for reading and writing behavior.}
\label{class_overlap}
\end{figure}

During the iterative training process on SCB-Dataset3-S, we encountered an issue where there was overlap between the bounding boxes of the "reading" and "writing" behaviors in the behavior detection (see Fig.~\ref{class_overlap}). Specifically, there were 8 instances where the bounding boxes overlapped between these two behaviors. This observation indirectly suggests that "reading" and "writing" behaviors are visually similar, making it more challenging for the YOLOv7 model to distinguish between them. To improve the accuracy of the model, further exploration is needed to address this challenge.

\begin{figure}
\centerline{\includegraphics[width=0.48\textwidth]{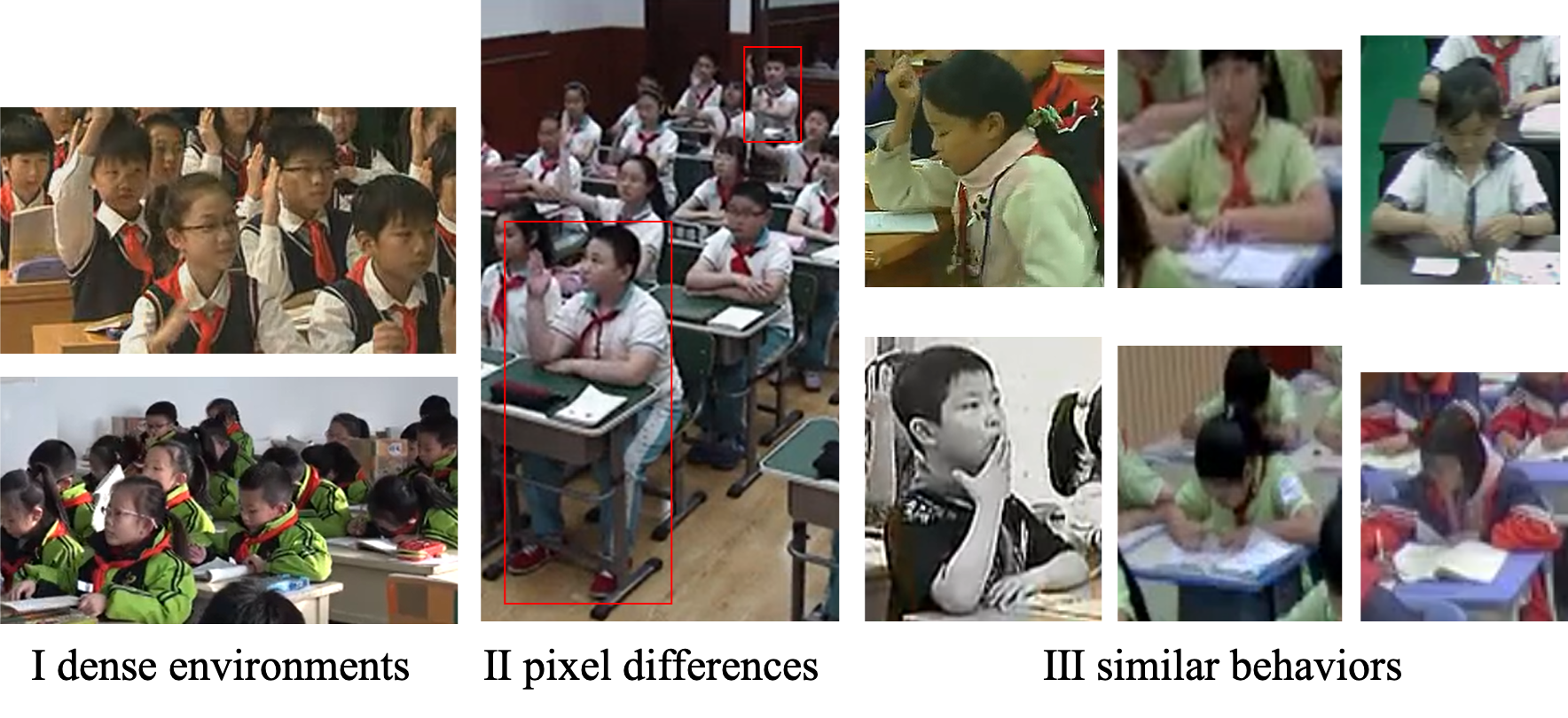}}
\caption{Challenges in the SCB-Dataset3-S include dense environments, similar behaviors, and pixel differences..}
\label{dense_pixel_similar}
\end{figure}

And the classroom environment presents challenges for detecting behavior due to the crowded students and variation in positions, as shown in Fig.~\ref{dense_pixel_similar} \uppercase\expandafter{\romannumeral1} and \uppercase\expandafter{\romannumeral2}. In addition to the high degree of similarity between reading and writing in certain scenarios, there is also a visual similarity between hand-raising and other classes of behavior. This further complicates the detection process, as shown in Fig.~\ref{dense_pixel_similar} \uppercase\expandafter{\romannumeral3}.

\begin{figure}
\centerline{\includegraphics[width=0.48\textwidth]{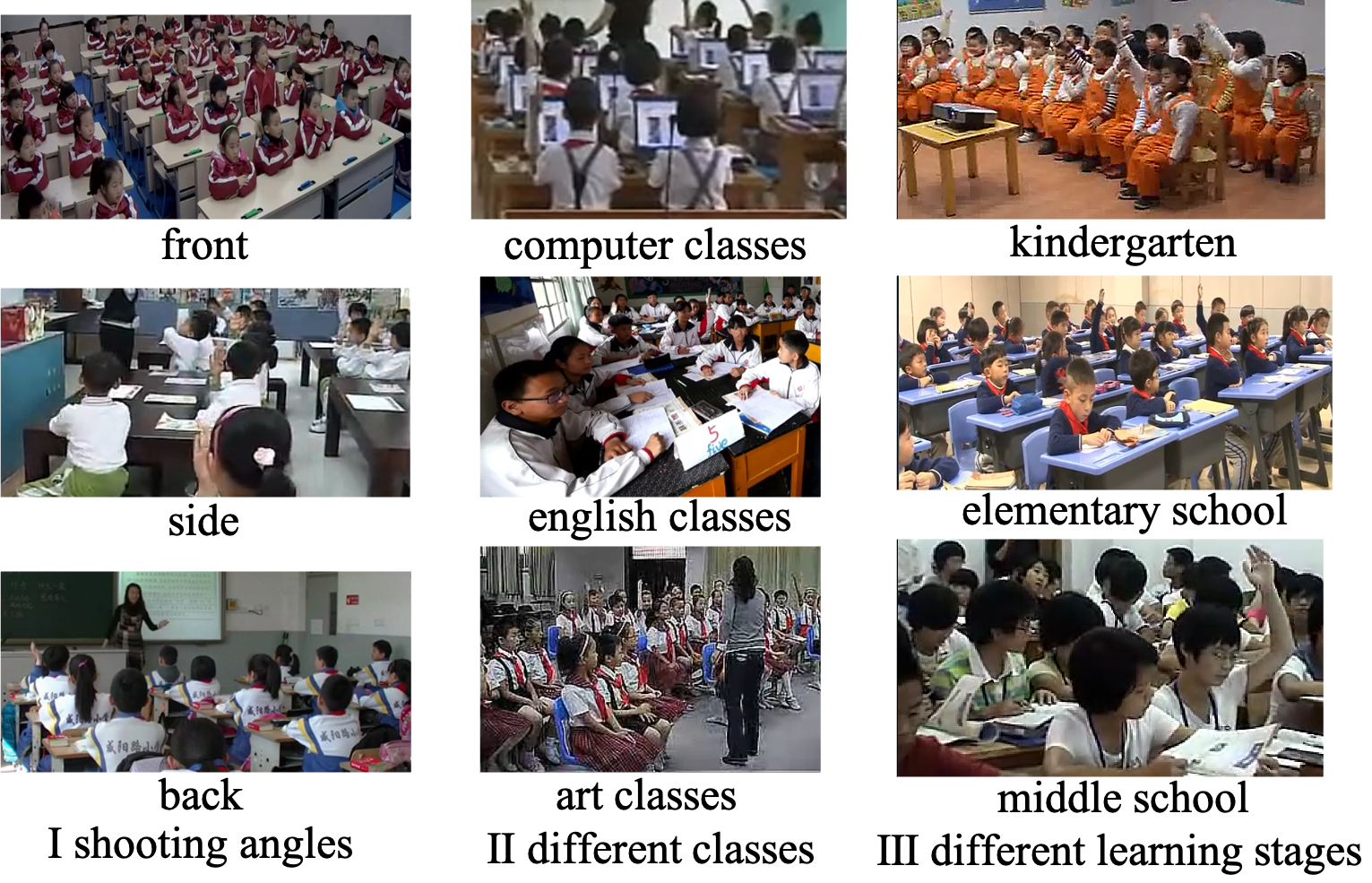}}
\caption{Challenges in the SCB-Dataset3-S include varying shooting angles, class differences, and different learning stages.}
\label{angles_class_stages}
\end{figure}

The SCB-Dataset3-U was collected from different angles, including front, side, and back views (Fig.~\ref{angles_class_stages} \uppercase\expandafter{\romannumeral1}). Additionally, the classroom environment and seating arrangement can vary, which adds complexity to the detection task (Fig.~\ref{angles_class_stages} \uppercase\expandafter{\romannumeral2}). Student classroom behavior behaviors also differ across learning stages from kindergarten to high-school, creating challenges for detection (Fig.~\ref{angles_class_stages} \uppercase\expandafter{\romannumeral3}).

However, for university classroom data, the situation is different from kindergarten to high school data. In online open courses, there are rarely real videos of students in the classroom, and even if there are, they are usually staged. The focus of these videos is mainly on the teacher's course content, rather than the students' classroom performance. Obviously, this does not meet the needs of our dataset. Therefore, in order to focus on real classroom learning behaviors of college students, we collected real classroom recordings from university. Student classroom behavior dataset in university called SCB3-Dataset-U.

\begin{figure}[htbp]
\centerline{\includegraphics[width=0.48\textwidth]{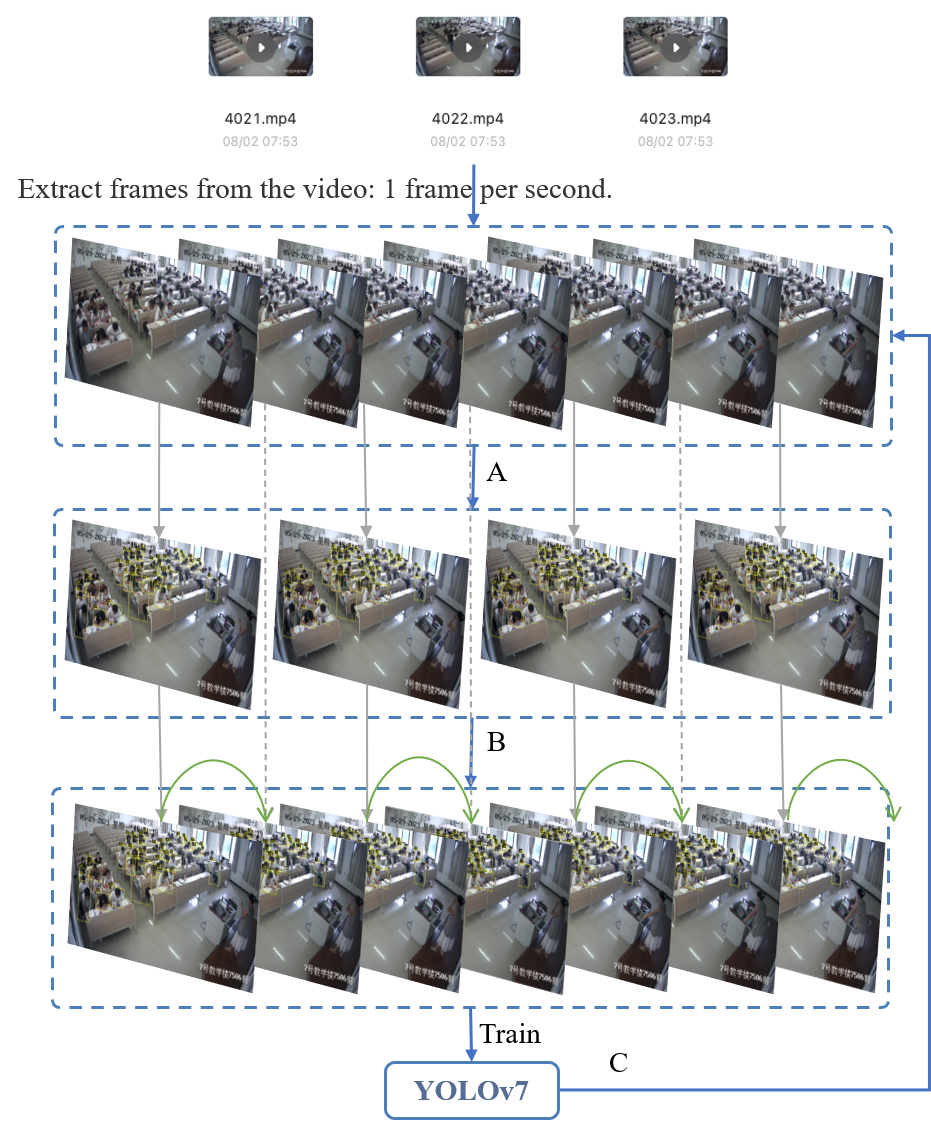}}
\caption{Process for creating SCB-Dataset3-U.}
\label{Process-of-SCB3-U}
\end{figure}

Due to the limited availability of suitable videos, the data processing method for SCB-Dataset3-U is different from the dataset for kindergarten to high school. As shown in Fig.~\ref{Process-of-SCB3-U} , the production process of SCB-Dataset3-U uses the "frame interpolation data" \cite{ava-kinetics} and "iterative training" methods. Firstly, we extract one frame image every 1 second from the videos. Then, we annotate one image every 2 second (as shown in Fig.~\ref{Process-of-SCB3-U} A). In our experiments, we found that annotating every 1 to 5 seconds is sufficient to capture the specific behaviors of the students. Considering factors such as data volume, error rate, and labor cost, we choose to annotate every 2 second. Next, we pass the annotation data of the current frame to the next frame (as shown in Fig.~\ref{Process-of-SCB3-U} B). Finally, we use the annotated data to train the YOLOv7 network and expand the dataset using iterative training (as shown in Fig.~\ref{Process-of-SCB3-U} C).

Our SCB-Dataset3-U dataset annotations focus only on the behavior of the current frame and do not include annotations for the relationships between frames. This approach has certain advantages. On one hand, it helps to reduce manpower costs. On the other hand, it allows for the construction of an image dataset, which serves as a foundation for future understanding of videos. Once the relationships between these frames are determined, a complete spatio-temporal behavior video dataset can be formed.

\begin{table}
    \centering \renewcommand\arraystretch{1.25} 
    \caption{comparisons of SCB-Dataset1, SCB-Dataset2, SCB-Dataset3-S, SCB-Dataset3-U.}
    \label{comparisons_SCB123}
    \begin{tabular}{ccccc}
        \hline
         & SCB1 & SCB2 & SCB3-S & SCB3-U \\ \hline
        classes & 1 & 3  & 3 & 6 \\
        Images & 4001 & 4266 & 5015 & 671 \\
        Annotations & 11248 & 18499 & 25810 & 19768 \\
        \hline
    \end{tabular}
\end{table}

The Table  \ref{comparisons_SCB123} shows the one-year iteration process of our student behavior dataset, where SCB1 represents SCB-Dataset1, SCB2 represents SCB-Dataset2, SCB3-S represents SCB-Dataset3-S, and SCB3-U represents SCB-Dataset3-U. The "class" row indicates the behaviors. The number 1 represents the hand-raising behavior, the number 3 represents the hand-raising, reading, and writing behaviors, and the number 6 represents the hand-raising, reading, writing, using phone, bowing the head, and leaning over the table behaviors. It can be seen from the table that our dataset has increased in both class and image quantity, with a particularly noticeable increase in the number of annotations. By comparing the number of annotations and the number of images, it can be observed that although SCB-Dataset3-U has a lower number of images, the average number of annotations per image is the highest.

\begin{table}
    \centering \renewcommand\arraystretch{1.25} 
    \caption{Statistics of behavior classes in different datasets.}
    \label{Statistics_behavior_classes}
    \begin{tabular}{ccccccc}
        \hline
         & \makecell{hand-\\raising} & reading & writing & \makecell{using \\phone} & \makecell{bowing \\the head} & \makecell{leaning over \\the table}  \\ \hline
        SCB1 & 11248 & - & - & - & - & - \\
        SCB2 & 10078 & 5882 & 2539 & - & - & - \\
        SCB3-S & 11207 & 10841 & 3762 & - & - & - \\
        SCB3-U & 6 & 7826 & 2984 & 6976 & 947 & 1029 \\
        \hline
    \end{tabular}
\end{table}

Table  \ref{Statistics_behavior_classes} displays the number of different behavior classes in each dataset. It can be observed that the number of the "hand-raising" behavior class did not increase, but rather slightly decreased. During the dataset iteration process, we deliberately excluded low-quality images and tried to address the issue of imbalanced class data. Hence, in the later stages, we hardly collected any additional data for the "hand-raising" behavior. However, despite our deliberate selection process, class imbalance still persists, as can be seen from the data in Table 2. It is worth noting that there are very few instances of "using a mobile phone" and "resting on the desk" behaviors in the datasets from kindergarten to high school. Conversely, in the university dataset, there are fewer instances of the "hand-raising" behavior.

\begin{figure}[htbp]
\centerline{\includegraphics[width=0.48\textwidth]{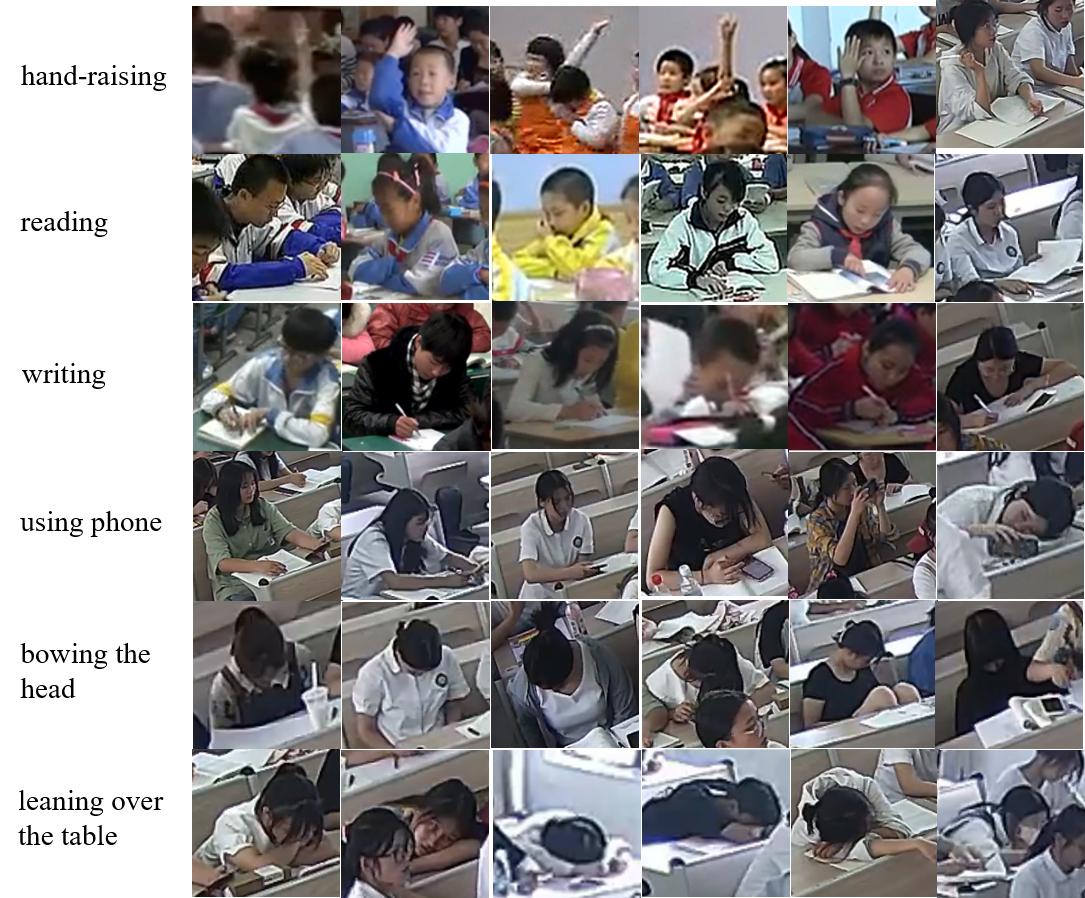}}
\caption{Example images of classroom behavior dataset.}
\label{Example_images_of_classroom_behavior_dataset}
\end{figure}

Some visual examples of the SCB-Dataset3 are illustrated in Fig.~\ref{Example_images_of_classroom_behavior_dataset}

\section{Behavioral similarity analysis}

\section{Experiments}

\subsection{Experimental Environment and Dataset}

The experiment was compiled and tested using python 3.8, the corresponding development tool was PyCharm, the main computer vision library was python-OpenCV 4.1.2, the deep learning framework used was Pytorch v1.11 with CUDA version 11.3 for model training, the operating system was Ubuntu 20.04.2, and the CPU: 12 vCPU Intel(R) Xeon(R) Platinum 8255C CPU @ 2.50GHz.  The GPU: RTX 3080(10GB) * 1.

The dataset used in our experiments is SCB-Dataset3-S and SCB-Dataset3-U, which we split into training, validation sets with a ratio of 4:1.

\subsection{Model Training}

To train the model, set the epoch to 100, batch size to 8, and image size to 640x640, and we use a pre-trained model for the training.

\subsection{Evaluation Metric}

When evaluating the results of an experiment, we use two main criteria: Precision and Recall. True Positive (TP) represents a correct identification, False Positive (FP) means an incorrect identification, and False Negative (FN) indicates that the target was missed.

To calculate Precision (Eq. \ref{precision}), we divide the number of True Positives by the sum of True Positives and False Positives. For Recall (Eq. \ref{recall}), we divide the number of True Positives by the sum of True Positives and False Negatives. Both Precision and Recall must be considered to properly assess the accuracy of the experiment.

\begin{equation}
    \centering
    \begin{aligned}
        precision~ = ~\frac{TP}{TP + FP}
    \end{aligned}
    \label{precision}
\end{equation}

\begin{equation}
    \centering
    \begin{aligned}
        recall~ = ~\frac{TP}{TP + FN}
    \end{aligned}
    \label{recall}
\end{equation}

 To provide a more thorough evaluation of Precision and Recall, the metrics of Average Precision (AP, Eq. \ref{AP}) and mean Average Precision (mAP, Eq. \ref{mAP}) have been introduced.  These metrics calculate the average Precision over a range of Recall values, which provides a more comprehensive assessment of the model's performance.  AP is the average of Precision values at all Recall levels, and mAP is the mean AP value averaged over different classes.  
 
 \begin{equation}
    \centering
    \begin{aligned}
        AP_{i} = {\int_{0}^{1}{P(r)dr}}\\
    \end{aligned}
    \label{AP}
\end{equation} 

\begin{equation}
    \centering
    \begin{aligned}
        mAP = \frac{1}{n}{\sum\limits_{i}^{n}\left( AP_{i} \right)}
    \end{aligned}
    \label{mAP}
\end{equation}

As shown in Table \ref{comparison_of_YOLO_on_SCB-Dataset3-S}, mAP@50 refers to the mean average precision at IoU threshold 0.5, while mAP@50:95 represents the mean average precision at IoU thresholds ranging from 0.5 to 0.95.

\begin{equation}
    \centering
    \begin{aligned}
    BSI = ( \frac{S_{i,j}}{N_{i}},  \frac{S_{i,j}}{N_{j}})
    \end{aligned}
    \label{BSI}
\end{equation}

To calculate the similarity of student behaviors, we introduce the BSI (Behavior Similarity Index), as shown in Eq. \ref{BSI}. In the equation, $S_{ij}$ represents the overlap count between behavior class i and behavior class j,  $N_{i}$ represents the count of behavior class i, and $N_{j}$ represents the count of behavior class j. Therefore, BSI represents the overlap rate of behavior 7 classes i and j, and BSI can to some extent reflect the similarity between behaviors.

\subsection{Analysis of experimental results}

For our training on SCB-Dataset3-, we employed YOLOv5, YOLOv7 YOLOv8's diverse range of network structures, encompassing YOLOv5n, YOLOv5s, YOLOv5m, YOLOv5l, YOLOv5x, YOLOv7, YOLOv7x, YOLOv7w6, YOLOv8n, YOLOv8s, YOLOv8m, YOLOv8l, YOLOv8x.  The outcomes of our experiments are detailed in Table \ref{comparison_of_YOLO_on_SCB-Dataset3-S}, with precision represented by "p" and recall represented by "R".

\begin{table}
    \centering \renewcommand\arraystretch{1.25} 
    \caption{The comparison of YOLOv5, YOLOv7, YOLOv8 on SCB-Dataset3-S.}
    \label{comparison_of_YOLO_on_SCB-Dataset3-S}
    \begin{tabular}{ccccccccc}
        \hline
         Models& P (\%) & R (\%) &  mAP@50 (\%) &  mAP@50:95(\%)   \\ \hline
        YOLOv5n & 68.1  & 67.5 & 71.1 & 48.3  \\
        YOLOv5s & 73.3 & 67.8 & 72.9 & 53.3 \\
        YOLOv5m & 74.4 & 70.0 & 74.7 & 58.2  \\
        YOLOv5l & 75.6 & 71.5 & 76.0 & 60.3   \\
        YOLOv5x & \underline{\textbf{75.8}} & 69.3 & 74.9 & 60.1  \\
        YOLOv7 & 75.0  & 73.6 & 77.2 & 60.7  \\
        YOLOv7x & 75.6  & \underline{\textbf{75.9}} & \underline{\textbf{80.3}} & \underline{\textbf{63.1}} \\
        YOLOv7w6 & 74.6 & 74.6 & 80.0 & 62.3 \\
        YOLOv8n & 67.9 & 67.7 & 72.4 & 55.0 \\
        YOLOv8s & 68.5 & 72.1 & 74.5 & 58.3 \\
        YOLOv8m & 72.7 & 74.0 & 77.6 & 61.7   \\
        YOLOv8l & 75.4 & 71.2 & 76.8 & 61.2  \\
        YOLOv8x & 74.8 & 71.2 & 76.4 & 61.7   \\
        \hline
    \end{tabular}
\end{table}

From Table \ref{comparison_of_YOLO_on_SCB-Dataset3-S}, it can be observed that YOLOv7x performs the best in terms of training and testing results for SCB-Datat3-S, achieving an mAP@50 of 80.3$\%$. Additionally, when considering the detection performance of YOLOv5, YOLOv7, and YOLOv8, it is evident that larger model sizes do not necessarily yield better results, the ranking of model sizes is as follows:

YOLOv5n$<$YOLOv5s$<$YOLOv5m$<$YOLOv5l$<$YOLOv5x

YOLOv7$<$YOLOv7x$<$YOLOv7w6

YOLOv8n$<$YOLOv8s$<$YOLOv8m$<$YOLOv8l$<$YOLOv8x.

\begin{table*}
    \centering \renewcommand\arraystretch{1.25} 
    \caption{Detection Results of Various classes in SCB-Dataset3-S using YOLOv5, YOLOv7, and YOLOv8}
    \label{Various_classes_on_YOLO}
    \begin{tabular}{ccccccccccccc}
        \hline
         \multirow{2}{*}{Models} & \multicolumn{4}{c}{hand-raising (\%)}   &  \multicolumn{4}{c}{reading (\%)} &  \multicolumn{4}{c}{writing (\%)}   \\  
        & P & R &  AP@50 &  AP@50:95& P & R &  AP@50 &  AP@50:95& P & R &  AP@50 &  AP@50:95 \\ \hline
        YOLOv5n & 79.4 & 72.7 & 80.6 & 51.9 & 66.5 & 70.0 & 72.9 & 51.4 & 58.4 & 59.7 & 59.8 & 41.6 \\
        YOLOv5s & 85.4 & 75.3 & 82.8 & 58.4 & 71.5 & 68.3 & 73.2 & 54.9 & 63.0 & 59.9 & 62.6 & 46.7 \\
        YOLOv5m & 86.0 & 78.8 & 85.5 & 64.2 & 72.0 & 70.9 & 75.3 & 60.0 & 65.2 & 60.2 & 63.4 & 50.6 \\
        YOLOv5l & 87.1 & 79.3 & 85.8 & 66.6 & 72.6 & 71.7 & 75.2 & 60.8 & 67.0 & 63.5 & 66.9 & 53. 4\\
        YOLOv5x & \underline{\textbf{88.2}} & 77.8 & 85.6 & \underline{\textbf{66.9}} & 72.8 & 70.1 & 74.6 & 61.2 & 66.4 & 60.0 & 64.5 & 52.2\\
        YOLOv7 & 86.2 & \underline{\textbf{81.6}} & 86.8 & 66.2 & 72.3 & 73.6 & 77.5 & 62.3 & 66.5 & 65.5 & 67.2 & 53.6\\
        YOLOv7x & 85.9 & 81.0 & 87.7 & 66.7 & 71.1 & \underline{\textbf{76.7}} & \underline{\textbf{79.9}} & \underline{\textbf{63.9}} & \underline{\textbf{69.7}} & \underline{\textbf{69.9}} & \underline{\textbf{73.2}} & \underline{\textbf{58.6}} \\
        YOLOv7-w6 & 84.9 & 80.7 & \underline{\textbf{87.9}} & 66.7  & 72.8 & 73.4 & 79.8 & 62.5 & 66.0 & 69.7 & 72.3 & 57.7  \\
        YOLOv8n & 79.5 & 74.3 & 81.9 & 60.6 & 68.4 & 70.0 & 74.8 & 57.7 & 55.8 & 58.6 & 60.6 & 46.7\\
        YOLOv8s & 80.1 & 78.4 & 83.7 & 64.1 & 67.5 & 74.5 & 76.3 & 60.6 & 57.9 & 63.3 & 63.4 & 50.4\\
        YOLOv8m & 83.7 & 80.5 & 85.9 & 66.2 & 71.2 & 73.9 & 77.6 & 62.7 & 63.1& 67.5 & 69.4 & 56.1\\
        YOLOv8l & 85.2 & 77.8 & 84.8 & 65.5 & \underline{\textbf{73.0}} & 73.0 & 76.8 & 62.6 & 68.0 & 62.7 & 68.6 & 55.6\\
        YOLOv8x & 84.7 & 80.1 & 85.1 & 67.2 & 72.6 & 71.8 & 75.8 & 62.1 & 67.2 & 61.6 & 68.3 & 55.7 \\
        \hline
    \end{tabular}
\end{table*}

From Table \ref{Various_classes_on_YOLO}, it can be observed that there is a certain correlation between the accuracy of each class and the number of classes. For behaviors with a larger number of categories, such as hand-raising, the accuracy is higher; while for behaviors with a smaller number of classes, such as writing, the accuracy is relatively lower. Therefore, in the subsequent data iteration, we will reduce the data imbalance. Additionally, it can be observed from Table \ref{Various_classes_on_YOLO} that the YOLOv7x model has generally higher accuracy compared to other metrics.

\begin{table}
    \centering \renewcommand\arraystretch{1.25} 
    \caption{Results of BSI for different network models on SCB-Dataset3-S}
    \label{different_network_BSI}
    \begin{tabular}{ccccccccc}
        \hline
         Models& BSI(0,1) & BSI(0,2) &  BSI(1,2)   \\ \hline
        YOLOv5n & ( 0.5, 0.4 )  & ( 0.1, 0.2 ) & ( 3.8, 14.1 )  \\
        YOLOv5s & ( 0.5, 0.4 ) & ( 0.2, 0.5 ) & ( 3.3, 11.9 )  \\
        YOLOv5m & ( 0.4, 0.4 ) & ( 0.3, 0.8 ) & ( 3.2, 11.8 )  \\
        YOLOv5l & ( 0.4, 0.3 ) & ( 0.0, 0.1 ) & ( 3.3, 12.3 )   \\
        YOLOv5x & ( 0.5, 0.5 )  & ( 0.0, 0.1 )  & ( 3.6, 13.2 )  \\
        YOLOv7 &  ( 0.7, 0.6 ) &  ( 0.1, 0.3 ) & ( 5.7, 18.6 )  \\
        YOLOv7x & ( 0.7, 0.5 )  & ( 0.0, 0.1 ) & ( 4.9, 17.2 ) \\
        YOLOv7w6 & ( 0.3, 0.3 )  & ( 0.0, 0.1 ) & ( 6.0, 21.3 )  \\
        YOLOv8n & ( 0.3, 0.3 ) & ( 0.0, 0.1 ) & ( 4.5, 15.8 ) \\
        YOLOv8s & ( 0.3, 0.3 ) &  ( 0.0, 0.1 ) & ( 4.2, 15.5 ) \\
        YOLOv8m & ( 0.4, 0.3 ) & ( 0.1, 0.2 ) & ( 3.6, 12.9 )   \\
        YOLOv8l & ( 0.4, 0.3 ) & ( 0.2, 0.3 ) & ( 3.1, 11.8 )  \\
        YOLOv8x & ( 0.1, 0.1 ) & ( 0.0, 0.0 ) & ( 2.4, 9.2 )   \\
        \hline
    \end{tabular}
\end{table}

In Table \ref{different_network_BSI}, we used 0 to represent hand-raising, 1 to represent reading, and 2 to represent writing. Based on the results from Table \ref{different_network_BSI}, we can observe that hand-raising behavior has almost no similarity with other behaviors, while reading and writing exhibit a higher degree of similarity.

\begin{table}
    \centering \renewcommand\arraystretch{1.25} 
    \caption{SCB-Dataset3-U's YOLOv7x training and testing results (prior to "frame interpolation data" processing).}
    \label{SCBDataset3-U_YOLOv7x_prior}
    \begin{tabular}{ccccc}
        \hline
         \multirow{2}{*}{\shortstack{Class}} & \multirow{2}{*}{\shortstack{P($\%$)}} & \multirow{2}{*}{\shortstack{R($\%$)}} & \multirow{2}{*}{\shortstack{mAP@\\50($\%$)}}  & \multirow{2}{*}{\shortstack{mAP@\\50:95($\%$)}}   \\  \\ \hline
        all & 91.1  & 57.0 & 71.4 & 55.1 \\
        hand-raising & 100.0 & 0.0 & 0.0 & 0.0  \\
        reading & 85.6 & 72.9 & 81.9 & 65.7  \\
        writing & 83.6 & 82.7 & 88.3 & 70.9  \\
        using phone & 87.4  & 86.8  & 92.8  & 77.3 \\
        bowing the head &  92.2 &  50.1 & 74.8 & 54.3 \\
        \multirow{2}{*}{\shortstack{leaning over \\ the table}} & \multirow{2}{*}{98.0} & \multirow{2}{*}{49.7} & \multirow{2}{*}{90.3} & \multirow{2}{*}{62.1} \\
        & & & \\
        \hline
    \end{tabular}
\end{table}

\begin{table}
    \centering \renewcommand\arraystretch{1.25} 
    \caption{SCB-Dataset3-U's YOLOv7x training and testing results (after "frame interpolation data" processing).}
    \label{SCBDataset3-U_YOLOv7x_after}
    \begin{tabular}{ccccc}
        \hline
         \multirow{2}{*}{\shortstack{Class}} & \multirow{2}{*}{\shortstack{P($\%$)}} & \multirow{2}{*}{\shortstack{R($\%$)}} & \multirow{2}{*}{\shortstack{mAP@\\50($\%$)}}  & \multirow{2}{*}{\shortstack{mAP@\\50:95($\%$)}}   \\  \\ \hline
        all & 92.9  & 75.0 & 94.4 &  74.1 \\
        hand-raising & 100.0 & 0.0 & 99.5 & 69.7  \\
        reading & 88.4 & 85.1 & 92.1 & 74.6  \\
        writing & 89.6 & 85.9 & 92.3 & 78.0  \\
        using phone & 93.6  & 94.9  & 96.9  & 80.6 \\
        bowing the head &  88.1 &  85.9 & 87.3 & 66.1 \\
        \multirow{2}{*}{\shortstack{leaning over \\ the table}} & \multirow{2}{*}{97.7} & \multirow{2}{*}{98.2} & \multirow{2}{*}{98.5} & \multirow{2}{*}{75.5} \\
        & & & \\
        \hline
    \end{tabular}
\end{table}

For SCB-Dataset3-U, due to its limited data quantity, we only conducted training and testing on YOLOv7x. We evaluated the accuracy before and after applying "frame interpolation" and the results are presented in Table \ref{SCBDataset3-U_YOLOv7x_prior} and \ref{SCBDataset3-U_YOLOv7x_after}.

From Table \ref{SCBDataset3-U_YOLOv7x_prior} and \ref{SCBDataset3-U_YOLOv7x_after}, it can be seen that the accuracy of SCB-Dataset3-U significantly improved after "frame interpolation" processing. However, there is still room for improvement in SCB-Dataset3-U at present. In this stage, we only used data from one class and one session, resulting in very similar training and testing data (same individuals and scenarios). This leads to inflated test results, and if we were to test in different scenarios, the performance is likely to decrease significantly. Therefore, in future work, we will increase the amount of data in SCB-Dataset3-U to ensure diversity of the data. Nevertheless, the current work has to some extent validated the effectiveness of "frame interpolation", as it can increase the amount of data, improve detection accuracy, and reduce manual labor costs.

\subsection{Conclusion}

In summary, this paper has made contributions to the field of student behavior detection in education.     Through the development of the SCB-dataset3 and its evaluation using the YOLOv5, YOLOv7, YOLOv8 algorithm, we have addressed the gaps in student data sets within the field and provided fundamental data for future research.     These contributions have the potential to enhance the accuracy and effectiveness of student behavior detection systems, ultimately benefiting both students and educators.

However, we acknowledge that the current classes and quantity of student behavior data sets are not sufficient, especially SCB-Dataset3-U. We plan to continuously update and improve SCB-Dataset3 and explore implementing a spatio-temporal behavior dataset for student classrooms. At the same time, we will further study how to solve the problem of low accuracy caused by similar behaviors. Additionally, we will also explore more neural networks, especially spatio-temporal behavior networks. We believe that our work will help to advance the application of artificial intelligence in education, enhance teaching effectiveness, and ultimately benefit students.     Our SCB-dataset can be downloaded from the link provided in the abstract for further research and development.

\FloatBarrier

\end{document}